\documentclass[11pt,a4paper]{article}
\pdfoutput=1

\usepackage[hyperref]{acl2019}
\usepackage{times}  
\usepackage{latexsym}
\usepackage{url}  
\usepackage{graphicx}  
\usepackage{amsmath}
\usepackage{float}

\aclfinalcopy 
\def\aclpaperid{61} 

\newcommand{\eat}[1]{}
\newcommand{\sys}{HEIDL}

\title{\sys: Learning Linguistic Expressions with Deep Learning\\ and Human-in-the-Loop} 

\author{
Yiwei Yang\textsuperscript{1},
Eser Kandogan\textsuperscript{2},
Yunyao Li\textsuperscript{2},
Walter S. Lasecki\textsuperscript{1},
Prithviraj Sen\textsuperscript{2}\\
\textsuperscript{1}Computer Science and Engineering, University of Michigan - Ann Arbor\\
\textsuperscript{2}IBM Research - Almaden, San Jose, CA\\
\{yanyiwei, wlasecki\}@umich.edu,
\{eser, yunyaoli, senp\}@us.ibm.com
}

\date{}

\begin{document}

\maketitle

\begin{abstract}
While the role of humans is increasingly recognized in machine learning community, representation of and interaction with models in current human-in-the-loop machine learning (HITL-ML) approaches are too low-level and far-removed from human's conceptual models.  We demonstrate \sys, a prototype HITL-ML system that exposes the machine-learned model through high-level, explainable linguistic expressions formed of predicates representing semantic structure of text. In \sys, human's role is elevated from simply evaluating model predictions to interpreting and even updating the model logic \textit{directly} by enabling interaction with rule predicates themselves. Raising the currency of interaction to such semantic levels calls for new interaction paradigms between humans and machines that result in improved productivity for text analytics model development process. Moreover, by involving humans in the process, the human-machine co-created models generalize better to unseen data as domain experts are able to instill their expertise by extrapolating from what has been learned by automated algorithms from few labelled data. 
\end{abstract}

\section{Introduction}
\label{intro}

Machine learning (ML) is an inherently iterative process where humans, ML experts, play a central role. Experts decide which features to include, hyperparameters to tune, metrics to evaluate, and whether the desired level of quality has been attained, failing which they iterate all over. Traditionally, to understand a predictive model one usually begins by examining the model's predictions with little understanding of the inner workings of the learned, \emph{black-box} model. 
More transparent representations of predictive models, such as first-order logic (a dialect with human-interpretable semantics), allows understanding the inner workings of the learned model, but traditional techniques~\citep{muggleton:ilpw96} to learning these are too brittle for real-world data, as they fail to learn anything unless there exists a logic program that can perfectly separate the data according to its labels. Deep learning has been used to successfully learn rule-based predictive models~\citep{cohen:arxiv17,yang:nips17,evans:jair18} from data with noisy labels. Besides interpretability and explainability, other advantages of learning a logical model include the promise of improved generalization to unseen data due to the strong inductive bias provided by the predicates employed~\citep{evans:jair18}.

Human-in-the-loop machine learning (HITL-ML) approaches aim to provide humans with the ability to interact with the model that goes beyond simply examining model predictions. Humans need to be able to interpret, explain, and reason about models throughout the model development cycle, especially in domains where labeled training data is too limited to learn a model that generalizes well to unseen data, we need to be able to interpret and examine machine learning models. The above-mentioned works that utilize deep learning to learn a logical theory raise new opportunities and challenges since their precise syntax is more readily interpretable by humans that allows for new ways for humans to interact with machine learned models at levels that go far beyond just inspecting predictions. 

In this demo, we describe a HITL-ML approach for text analytics that exposes the machine-learned model through abstract, semantic, explainable rules, and allows humans and domain experts to examine, interact, and even modify the model logic directly. We present \emph{\sys} (\underline{H}uman-in-the-loop linguistic \underline{E}xpressions w\underline{I}th \underline{D}eep \underline{L}earning), a system designed to help domain experts access the linguistic expressions or rules learned with deep learning for text analytics, inspect them, show their inner workings by exposing how they operate on examples, and even breaking them apart into their constituent predicates and adding new ones to create new expressions in the process. \sys\ enables domain experts to instill their expertise into a machine-learned model thus resulting in a co-created one that has superior generalization performance than what is achievable by humans or machine learning optimization algorithms alone while incurring a fraction of human labor thus increasing the productivity of the overall text analytics model development process.

To evaluate \sys's efficacy, we conducted a user study where IBM\textsuperscript{\textregistered}'s data scientists used \sys\ to improve linguistic expressions for classifying sentences extracted from real-world, legal contracts. Since contracts may be proprietary, the initial learned linguistic expressions fed as input to \sys\ were trained on IBM\textsuperscript{\textregistered} procurement contracts while the test set consisted of non-IBM\textsuperscript{\textregistered} procurement contract sentences. Within $30$ minutes, each data scientist produced linguistic expressions with training set precision, recall upwards of $75\%$ representing a significant reduction in model development time since trawling through $>28,000$ training sentences to construct linguistic expressions from scratch would require multiple weeks of effort. In terms of prediction quality, while a black-box long short-term memory network trained by replacing tokens with GloVe embeddings~\citep{pennington:emnlp14} produces $44\%$ F1 on held out non-IBM\textsuperscript{\textregistered} procurement contract sentences, the linguistic expressions modified by data scientists via \sys\ produces $55\%$ F1 translating to a $25\%$ improvement in out-of-domain generalization performance. We emphasize that \sys\ is not specific to the task described here and may be useful for learning linguistic expressions for any classification task. Moreover, the main ideas embodied in \sys\ may be helpful for building tools to learn explainable models for applications beyond classification.

\section{Related Work}

Human computation integrates human effort into computational processes to complete tasks that cannot yet be done by computer. Prior work has focused on using crowdsourcing to facilitate and scale human computation. For instance, crowd-powered systems have been developed to support continuous conversations~\citep{chorus}, edit rules governing UI behaviors described in natural language~\citep{sketchexpress}, support natural language editing of papers~\citep{soylent}, and create novel stories~\citep{ensemble}. There is also work on evaluating task design tradeoffs on crowdsourced text annotations\citep{snow} and \citep{paraphrase}. Our approach draws on human computation by leveraging domain experts to create a more generalizable model even when data is scarce. 

Prior work in HITL-ML has focused on eliciting knowledge from people to create more powerful models. For example, research has introduced ways to solicit examples from people for labels to strengthen the training data, e.g., through active learning~\cite{settles2008analysis} queries that select the most informative item from an oracle to reduce labeling effort. In SEER~\citep{hanafi:sigmod17}, people select few examples for the system to learn information extraction rules. In AnchorViz~\citep{anchorviz}, people explore semantically related examples to find feature blinded items. Other work has focused on interactive clustering where people guide the model by changing the clusters~\citep{closingloop} and providing keywords~\citep{interactiveclustering}, and on eliciting feedback from experts to select relevant features~\citep{featurerelevance}. We also elicit user input to improve a model, but our approach presents understandable output, allowing users to directly modify it by selecting trusted rules.

\section{Approach}

\begin{figure}
\includegraphics[width=\columnwidth]{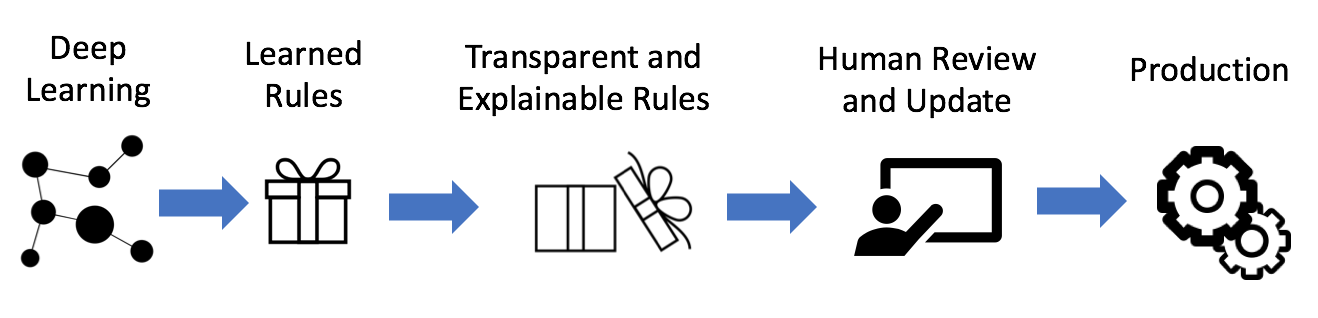}
\vspace{-1.8pc}
\caption{Overview of our approach}
\vspace{-0.8pc}
\label{fig:approach}
\end{figure}

Fig.\ref{fig:approach} shows an overview of our approach where we begin by learning linguistic expressions from labeled data using deep learning followed by using our system to explain said expressions to domain experts so they can verify and modify these. We briefly explain how to derive predicates from semantic linguistic structures, then describe how deep learning is used to construct linguistic expressions before describing our system.

\begin{figure*}[th!]
\centering
\includegraphics[width=.85\textwidth]{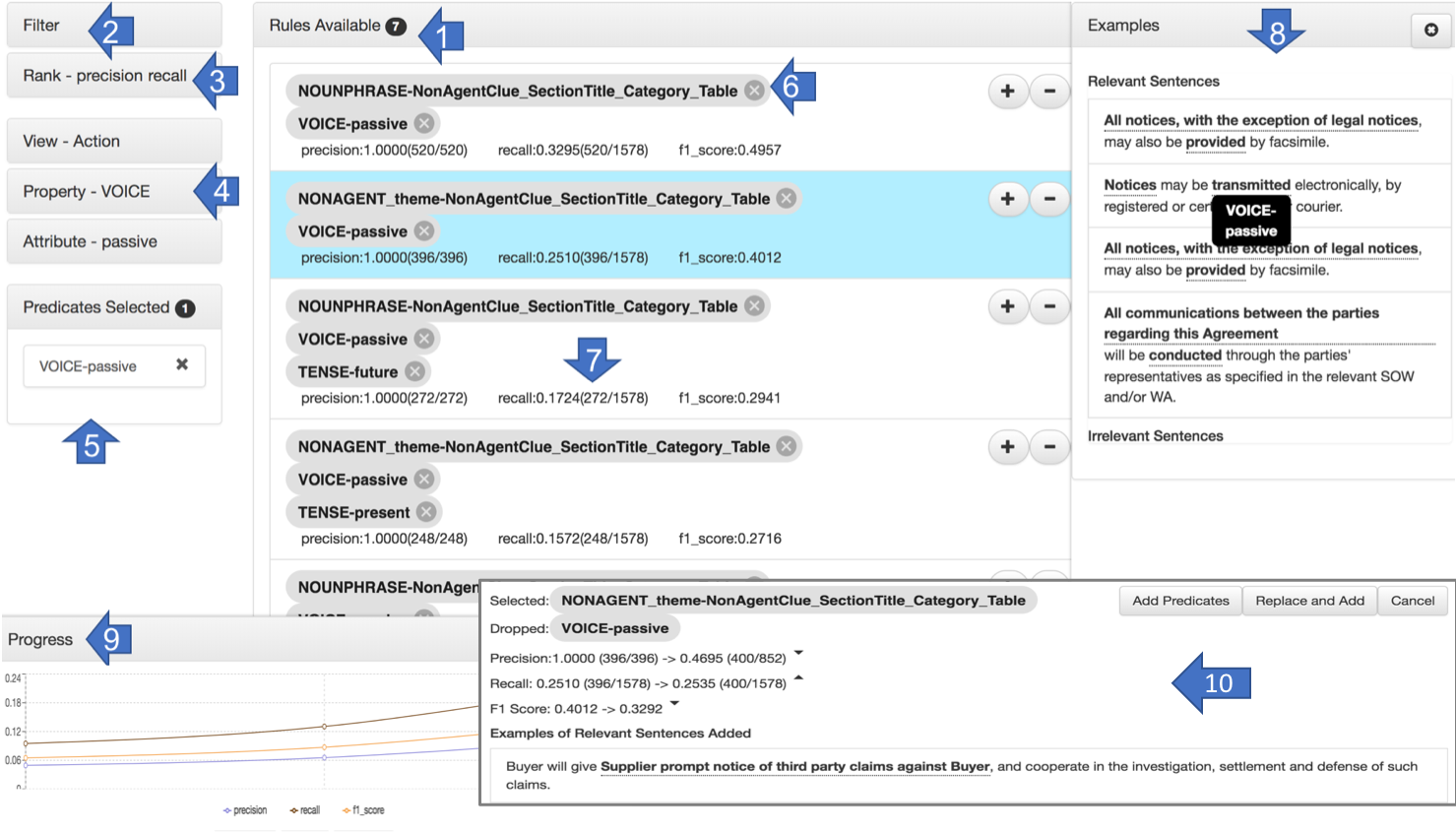}
\vspace{-.8pc}
\caption{UI allows users (1) to get an overview of linguistic expressions (2) filter by precision, recall, and F1, (3) rank, (4,5) filter by predicates, (6) remove expressions by predicate, (7) examine metrics and (8) examples for each expression, (9) monitor overall progress as users add and remove expressions to their collection, and (10) provide a `playground' allowing users to examine and modify expressions.}
\label{fig:overview}
\vspace{-0.9pc}
\end{figure*}

\subsection{Semantic Linguistic Structure (SLS)}

Each SLS refers to the shallow semantic representations corresponding to each sentence and generated automatically with natural language processing techniques such as semantic role labeling and syntactic parsing.  It captures ``who is doing what to whom, when, where and how'' described in a sentence as depicted by the following example (simplified for readability). 
\vspace{-1.3pc}
\begin{center}
$\underbrace{\text{John }}_{\text{{\tt agent}}} \underbrace{\text{bought }}_{\text{{\tt action}}} \underbrace{\text{daisy }}_{\text{{\tt theme}}} \text{for } \!\!\!\!\!\!\!\underbrace{\text{Mary }}_{\text{{\tt beneficiary}}} \hspace{6pt}\underbrace{\text{ yesterday}}_{\text{{\tt context:temporal}}}.$
\end{center}
\vspace{.1pc}
SLSs may be used as predicates to form linguistic expressions. For sentence classification, we extract SLSs using SystemT's~\citep{krishnamurthy:sigmodrecord09} semantic role labeler including but not limited to actions/verbs and various arguments of the action such as agent (doer of the action), object of the action and manner in which it is performed. From these, we construct: 1) predicates that test properties of the action such as tense, aspect, mood, modalclass, voice and polarity, and 2) predicates generated by looking up the extracted verb (bases), agents, objects etc. in dictionaries. For the user study task, we have access to a multitude of hand-crafted dictionaries that contain surface forms for verb (bases), objects, themes etc. We emphasize that \sys\ is not specific to a particular set of predicates and can work with any set of predicates that allow us to learn initial, high quality linguistic expressions. Just like many other text analytics applications, and particularly for the domain of legal contracts, dictionary match predicates offer one way to achieve this.

\subsection{Deep Learning Linguistic Expressions}
\label{weightedrules}
As mentioned in Section \ref{intro}, recent works use neural networks to learn a logical theory. Two powerful and extremely general formulations that can learn from noisy labeled data virtually any kind of logic program, including linguistic expressions for classification, are TensorLog~\citep{cohen:arxiv17} and $\partial$ILP~\citep{evans:jair18}. Describing the accompanying learning algorithms is out of scope for this demo proposal but we describe the \emph{weighted rules} model representation to ground our system and better describe the input to \sys. Given predicates $\mathcal{P} = \{\text{pred}_1, \ldots \text{pred}_m\}$ and binary class labeled data $\mathcal{D} = \{(\mathbf{x}_1, y_1), \ldots (\mathbf{x}_n, y_n)\}$ such that each label $y_i \in \{0,1\}$ and $\mathbf{x}_i$ denotes a sentence, the weighted rules model associates a non-negative weight with each rule or linguistic expression\footnote{Weighted rules are also popular in statistical relational learning (e.g., Markov logic networks~\citep{richardson:ml06}) and for learning closed paths in knowledge graphs~\citep{yang:nips17}, the latter being based on TensorLog~\citep{cohen:arxiv17}.}:
\vspace{-.6pc}
\begin{eqnarray*}
&w_1:& \ell(\mathbf{x}) \leftarrow \text{pred}^1_1(\mathbf{x}) \wedge \ldots \text{pred}^1_{k_1}(\mathbf{x})\\
&&\vdots\\
&w_N:& \ell(\mathbf{x}) \leftarrow \text{pred}^N_1(\mathbf{x}) \wedge \ldots \text{pred}^N_{k_N}(\mathbf{x})
\end{eqnarray*}
\vspace{-.2pc}
Intuitively, if sentence $\mathbf{x}$ satisfies a linguistic expression then $\mathbf{x}$ inherits the corresponding weight; the higher the weight the greater the chance of the sentence being assigned the label, i.e. $\ell(\mathbf{x})$ is true. While learning a model based on logic provides strong inductive bias that can help regularize the learned model and better generalize to unseen data, due to limited labeled data in many real-world applications (e.g., enterprise settings where privacy and proprietary ownership restricts the size of the training set) the risk of overfitting is not completely eliminated. Moreover, the presence of weights can hamper explainability since humans are much better at interpreting logical expressions. Our goal is to take a weighted rules model such as described above and modify it to a fully explainable, more generalizable model that consists of a set of linguistic expressions following simple yet powerful disjunctive semantics, in other words, sentence $\mathbf{x}$ is assigned the label if \emph{any} of the expressions hold true for it, which is where \sys\ comes into the picture.

\subsection{User Experience}

Unlike classical HITL-ML,  our approach allows people to interact with machine-learned linguistic expressions and facilitate co-operative model development. There are two primary challenges: (1) present users with a quick overview of learned expressions; enable them to organize, order, and navigate expressions effectively, (2) help understand each expression's semantics and quality through examples and statistics; deepen understanding by providing a `playground' to verify and modify expressions while examining impact (see Fig.\ref{fig:overview}). Below we describe some of the user experience features in more detail to address these challenges.

\subsubsection{Overview, Rank, Filter}
\label{simpler}

Initially, the system presents all machine-learned linguistic expressions along with their precision, recall, and F1 measures (relevant for classification tasks). Users can rank and filter to organize the expressions to process them. Ranking allows users to quickly see the expressions with high performance on training data and is especially useful when the list of expressions is large. Filtering allows users to narrow down to a small set of similar expressions without being overwhelmed. Users can filter expressions by setting a minimum threshold on multiple performance measures. Users can also filter expressions by their constituent predicates. Filtering by predicate is useful when the users reckon an expression potentially generalizable, and would like to see similar expressions that share common predicates.

\subsubsection{Linguistic Expression Selection}

The end goal of the system is to create a collection of trusted linguistic expressions. To do so, after evaluating an expression, users would add it to the `approved' or `disapproved' collection. When an expression is approved, the combined performance of all approved expressions is recomputed.  This helps users to keep track of their overall progress. To help users assess expressions, \sys\ provides a random sample of up to 4 true positive and 4 false positive matching example sentences. Each example is decorated with annotations highlighting the constituent predicates that form part of the expression when the cursor hovers on it (Fig.~\ref{fig:overview}). 

While \sys\ shows the performance of each linguistic expression individually, they may be misleading as the sentences retrieved by the expression may be already covered by other expressions that have already been approved. \sys\ provides a `look ahead' feature to see the potential `delta' effect of adding the expression to the approved collection so users can see if approving the expression would be beneficial, in which case they can take a closer look by reading its associated examples, or examine it in `Playground' mode. 

\subsubsection{Playground}

\sys\ provides a Playground mode that allows users to inspect and modify linguistic expressions by adding or dropping predicates, and examine the effects. While playing with expressions, users can also examine the `delta' examples. If a predicate is dropped, then the expression becomes more general, thus retrieving more sentences than it previously did. Conversely, if a predicate is added fewer sentences are retrieved and \sys\ shows examples of the difference. This is  beneficial because it allows users to see the effect of individual predicates. Adding new predicates is especially useful if experts have a sense for which predicates are potentially good. Performance measures and examples are updated accordingly, helping users decide whether or not to keep the modification. 

\section{Salient Results from a User Study} 
\label{userstudy}

To evaluate \sys's efficacy, we conducted a user study among IBM\textsuperscript{\textregistered} data scientists with NLP expertise and knowledge of legal contracts. We recruited $4$ data scientists -- a relatively large number given most teams in industry include only $1$-$2$ data scientists. The task was to label sentences with \emph{Communication} which implies some form of communication between the two parties involved in the contract. The training data consists of $28,174$ sentences extracted from $149$ IBM\textsuperscript{\textregistered} procurement contracts and the held out, test data consists of $1259$ sentences extracted from $11$ non-IBM\textsuperscript{\textregistered} procurement contracts. The initial set of $188$ weighted linguistic expressions learned using deep learning performed at $67\%$ F1 (the harmonic mean of precision and recall) on the test set. Note that, as mentioned in Section \ref{weightedrules}, weights can lead to lack of explainability. Thus the data scientist's task is to use \sys\ to generate from this initial set of linguistic expressions, a smaller set expressed in pure first-order logic that achieves maximal performance on the out-of-domain test set. For each data scientist, we initialize \sys\ with the initial linguistic expressions, the training set sentences, and the corresponding predicates and dictionaries. Our baseline is a well established sentence classification neural network, based on bi-directional long short-term memory (LSTM). More precisely, the LSTM replaces the tokens in the input sentence with their corresponding 300-dimensional GloVe embedding, computes an intermediate hidden state which are then max-pooled, fed into a fully connected layer and ReLU activation before passing it through sigmoid activation to get a probability of predicting the label. To improve training of the LSTM, we employed a variety of dropout regularizations: variational dropout after the embedding lookup layer, weight dropout in the LSTM layer, and dropout in the fully connected layer.

Each user took roughly $30$ minutes to build a model that performed to their satisfaction on the training set (recall that, data scientists did not have access to the held out test set) which is far less effort than what would be required if writing linguistic expressions from scratch (by our estimates, person-weeks to cover the close $30,000$ sentences in the training set). Moreover, users selected fewer expressions compared to deep learning's weighted rules ($5$-$8$ vs. $188$) indicating that \sys\ helps learn parsimonious models thus aiding explainability. Most importantly, \sys\ improves generalization. We did a post-hoc analysis, measuring F1 for all combinations of sets of rules learned by the 4 participants. We report the averages and standard deviations respectively for teams sizes: 1: .32(SD=.17), 2: .46(SD=.1), 3: .52(SD=.06), and 4: .55(SD=0), showing that F1 increases with team size. On average, teams of 3 produced a F1 of .52 on the test set, as compared to .44 produced by the LSTM, which lacks explainability. We can potentially create a model with less human effort if high expertise exists (i.e., one of our participants subsumed the others in all combinations), if we can determine expertise beforehand. We also chose the top-K best rules from the initial set of linguistic expressions (where K was determined by optimizing over the training set) which produced .41 on the test set. These results indicate that using \sys, data scientists can instill their domain expertise into learned linguistic expressions and achieve superior out-of-domain generalization than using supervised learning alone while incurring far less human effort than writing rules from scratch.

\section{Demo Overview}

Besides \emph{Communication}, other useful labels for labeling sentences in legal contracts include \emph{Term \& Termination} (sentences that state contract termination clauses) and \emph{Payment Terms} (sentences that mention payments). The demo will provide an opportunity for attendees to use \sys\ to solve such classification tasks by developing binary classifiers in the form of linguistic expressions. Attendees will be able to access the sentences extracted from our training set and explore the initial $188$ linguistic expressions learned using deep learning formed out of $183$ predicates along with their associated dictionaries to get a feel for what it takes for a data scientist to develop linguistic expressions for such a challenging real-world domain. In our experience, the first good linguistic expressions usually takes a few seconds of exploration to identify using \sys's ranking and filtering features. 
For example, consider the line: 
\emph{Notices may be {\bf transmit}ted electronically}

Clearly, this sentence is referring to exchanging notices between the two parties involved in the contract and hence communication. Such sentences can be labeled with \emph{Communication} via dictionary matching predicates where the dictionary checks for relevant verb (bases) such as such as `notify' and `transmit'. More complex linguistic expressions may require combining multiple predicates and entering `Playground' mode or understanding how expressions work through examples shown by \sys. The accompanying video (\url{https://youtu.be/kicfGDMKu-w}) shows \sys\ in action.

\section{Conclusion}

Our taxonomy contains many more labels relevant to the contracts domain, each requiring development of linguistic expressions. Testimonies from data scientists confirm that for all labeling tasks, \sys\ was extremely useful. \sys\ is not specifically designed to work with a specific set of predicates nor with rules learned solely using deep learning. As long as the text analytics classification task provides a set of dictionary matching predicates from which a learner can produce accurate initial linguistic expressions, \sys\ can help domain experts instill further domain expertise into them. Even if the domain expert does not modify the linguistic expressions, tools such as \sys\ provide a useful control point that allows humans to verify the model before deploying it which should prove useful in enterprise settings. \sys's rule-centric interface provides an interesting counter-point to other HITL-ML methods such as active learning and SEER which are more example-centric. \sys\ is meant to improve productivity of domain experts and while we hope to perform further evaluations in future (on other tasks e.g., sentiment labeling), initial results have yielded promising results.

\bibliography{ref}
\bibliographystyle{acl_natbib}

\end{document}